\pdfoutput=1

\documentclass[11pt]{article}

\usepackage[final]{acl}

\usepackage{times}
\usepackage{latexsym}

\usepackage[T1]{fontenc}

\usepackage[utf8]{inputenc}

\usepackage{times}
\usepackage{latexsym}
\usepackage{booktabs}
\usepackage{amsmath}
\usepackage{float}
\usepackage{adjustbox}
\usepackage{longtable}
\usepackage{xcolor,colortbl}
\usepackage{diagbox} 
\usepackage{array,multirow}
\usepackage{graphicx} 
\usepackage{placeins} 
\usepackage{microtype} 
\usepackage{enumitem} 
\usepackage{amsmath, bm} 
\usepackage{comment} 
\usepackage{amsmath, bm} 
\usepackage{subcaption}
\usepackage[russian,ukrainian,english]{babel}
\usepackage{CJKutf8}

\usepackage{csquotes}

\usepackage{microtype}

\usepackage{inconsolata}

\title{Linear Cross-Lingual Mapping of Sentence Embeddings}

\author{Oleg Vasilyev, Fumika Isono, John Bohannon \\
  Primer Technologies Inc. \\
  San Francisco, California \\
  \texttt{{oleg,fumika.isono,john}@primer.ai}\\}

\begin{document}
\maketitle
\begin{abstract}
Semantics of a sentence is defined with much less ambiguity than semantics of a single word, and we assume that it should be better preserved by translation to another language. If multilingual sentence embeddings intend to represent sentence semantics, then the similarity between embeddings of any two sentences must be invariant with respect to translation. 
Based on this suggestion, we consider a simple linear cross-lingual mapping as a possible improvement of the multilingual embeddings. We also consider deviation from orthogonality conditions as a measure of deficiency of the embeddings.
\end{abstract}

\section{Introduction}\label{sec:Intro}
The approximately linear mapping between cross-lingual word embeddings in different languages is based on assumption that the word semantic meaning is conserved in a translation \cite{https://doi.org/10.48550/arxiv.1309.4168}. The linearity is only approximate because the corresponding words in different languages have different cultural background, different multiple meanings and different dependencies on context \cite{patra-etal-2019-bilingual, zhao-gilman-2020-non, Cao2020MultilingualAlignment, Peng2022Understanding}.
There are multiple patterns of polysemy, and the corresponding counts of word senses are different in different languages \cite{SRINIVASAN2015124, CASAS201919}. 

We expect, however, that a sentence has a less ambiguous meaning than a word, simply because the sentence context reduces ambiguity of each of its words. Indeed, in \cite{kang-etal-2024-translate} it is demonstrated that additional context helps to reduce disambiguation errors.
The idea that a sentence semantics should be better conserved in a translation was used in \cite{reimers-gurevych-2020-making}. 

In Appendix~\ref{app:ambiguity_loss} we provide simple examples illustrating the loss of word ambiguity in a sentence, and suggest that a good translation can preserve the residual ambiguity, if any. 
The examples show that if semantics of a sentence is somewhat changed in translation, then a better translation is possible. Unlike a lone word, which often has different sets of meaning in different languages, a sentence is not only less ambiguous but also allow differently phrased translations, among which there is usually at least one that fully preserves the semantics.

In order to explore the preservation of sentence semantics in translation, we consider here a linear mapping between multilingual embeddings in two languages. Unlike the removal of a language-specific bias in each language separately \cite{yang-etal-2021-simple, xie-etal-2022-discovering}, this mapping depends on both languages of interest and, while computationally cheap, may provide a better correspondence between the embeddings. Our contribution:
\begin{enumerate}[topsep=0pt,itemsep=-1ex,partopsep=1ex,parsep=1ex]
    \item We suggest simple and computationally light improvement of the correspondence of sentence embeddings between two languages. The 'sentence' can be one or several contiguous sentences. 
    \item For our evaluation we introduce a dataset based on wikipedia news.
    \item We demonstrate a non-orthogonality of the linear mapping between multilingual embeddings as an example and a measure of deficiency of a multilingual embedding model.
\end{enumerate}

\section{Cross-Lingual Linear Mapping}\label{sec:Linear}
Translation of a word can lose or add some of its meanings. But meaning of a sentence or of several contiguous sentences is better defined, and a good translation in most cases (except special idiomatic cases) should preserve the semantics (Appendix \ref{app:ambiguity_loss}). Embeddings of the translated sentences should be rigidly related to embeddings of the original sentences: the semantic similarities (or distances) between different embeddings should be preserved. In this section we assume that the 'sentence' is either a (not too short) sentence, or a larger segment of a text.

Suppose we have $N$ sentences, translated from language $L$ to language $L'$, and then embedded into a space of the same dimension $M$ in each of these languages: the embeddings $e_1, ... e_N$ in $L$ and the embeddings $e'_1, ... e'_N$ in $L'$.
If the measure of semantic similarity in both spaces is cosine, then we should expect that the normalized embeddings $e_i$ and $e'_i$ are related by rotation (orthonormal transform T): 
\begin{equation} \label{eq:rotation}
e' = Te
\end{equation}
with the orthogonality condition
\begin{equation} \label{eq:orthonormal}
\sum_i{T_{ij}T_{ik}} = \delta_{jk}
\end{equation}
where $i,j,k = 0, 1, ... , M-1$.

If semantic similarity is measured by euclidean distance, and the embeddings are not normalized, then we should allow the orthogonal transform to be accompanied by dilation and shift:
\begin{equation} \label{eq:orthonormal_dilation_shift}
e' = \alpha Te + b
\end{equation}
The above transformations should be observed if the translations preserved the semantics of the sentences, and if the embeddings represent the semantics correctly. 

In the following section we will allow any linear transformation $(A, b)$ between the embeddings in $L$ and $L'$: 
\begin{equation} \label{eq:linear}
\tilde{e} = Ae + b
\end{equation}
For our illustration here we created embeddings by one of SOTA aligned multilingual sentence-embedding model, on a set of translated sentences (Section \ref{ssec:Linear_Results}). We optimize the linear transformation on a set of embeddings, so that the mean squared distance between $\tilde{e}$ and $e'$ is minimal.

In the next section we consider the obtained linear transformation $(A, b)$ from two points of view:
\begin{enumerate}[topsep=0pt,itemsep=-1ex,partopsep=1ex,parsep=1ex]
    \item Replacement of the original embeddings $e$ by the transformed embeddings $\tilde{e}$ can serve as a fast and computationally cheap way to improve cross-lingual matching or clustering of a mix of texts of both languages. 
    \item We can observe how close is the optimized transformation $(A, b)$ to the 'ideal' relation eq.\ref{eq:orthonormal_dilation_shift}, and thus judge how good the embeddings are.
\end{enumerate}

\section{Observations}\label{sec:Observations}
\subsection{Data}\label{ssec:Linear_Data}
For obtaining the linear transformation eq.\ref{eq:linear} between embeddings, in Section \ref{ssec:Linear_Results} we use dataset Tatoeba\footnote{https://huggingface.co/datasets/tatoeba}. Tatoeba has $13$ languages with at least $100K$ sentences translated from English to the language. 
We consider performance of the obtained transformations on sentences and text segments of different style from multilingual WikiNews dataset\footnote{https://huggingface.co//datasets//Fumika//Wikinews-multilingual} which we created from real news (Appendix \ref{app:wikinews}). The samples have WikiNews articles in English as well as at least one other language, among 34 languages.

We will limit ourselves to six languages $L'$ that have a reasonable amount of data: at least $100K$ samples (of translations from $L$ to English) in Tatoeba, and at least $400$ samples in Wikinews (Appendix \ref{app:wikinews}): German ($de$), Spanish ($es$), French ($fr$), Italian ($it$), Portuguese ($pt$) and Russian ($ru$). 
Wikinews is used here for evaluation, in Section \ref{ssec:Linear_Results}, in two variations:
\begin{enumerate}[topsep=0pt,itemsep=-1ex,partopsep=1ex,parsep=1ex]
    \item \textit{WN}: Title of news article in English is paired with the same title in language $L'$.
    \item \textit{WN-text}: Title of news article in English is paired with the lower half of the text of the article in language $L'$. We selected the lower part in order to avoid easy lexical intersections of first phrases of the text with the title. (The article is split by whichever end of sentence is closer to the middle.)
\end{enumerate}
The evaluation on title-title pairs gives us a strong out-of-domain experience, and evaluation on title-text pairs provides a (more difficult) flavor of asymmetry in a multilingual search.
We also evaluate the obtained transformations on Flores dataset \cite{guzman-etal-2019-flores, goyal-etal-2022-flores, nllbteam2022language}\footnote{https://huggingface.co/datasets/facebook/flores}, and on a Tatoeba subset left aside from training.

\subsection{Evaluation}\label{ssec:Linear_Results}
We obtained the transformation $(A,b)$ (eq.\ref{eq:linear}) for each language $L' = de, es, fr, it, pt, ru$ by (1) obtaining embeddings $e$ for English sentences and embeddings $e'$ for the sentence translations to language $L'$, and (2) training a simple linear layer with bias, using embeddings $e_i$ as the inputs, and embeddings $e'_i$ as the labels, with the distance $|\tilde{e}_i - e'_i|$ serving as loss function. For each language, $10K$ embedding pairs were set aside for the testing, and $10K$ embedding pairs were set aside and used for validation during the training. We used state of the art embeddings paraphrase-multilingual-mpnet-base-v2 \cite{reimers-2019-sentence-bert}\footnote{https://huggingface.co/sentence-transformers/paraphrase-multilingual-mpnet-base-v2} for obtaining the embeddings $e$ and $e'$.

We can evaluate the benefit of replacing the original embeddings $e$ by the transformed embeddings $\tilde{e}$ in different ways. In Table \ref{tab:eval_tatoeba} we consider several examples: $dD$, $dC$, $fD$, $fC$ - defined below. 

\begin{table}[th!]
\centering
\begin{tabular}{@{}llccccc@{}}
\hline
{data}&{lang}&{$dD$}&{$dC$}&{$fD$}&{$fC$}\\
\hline
    \parbox[t]{2mm}{\multirow{5}{*}{\rotatebox[origin=c]{90}{Tatoeba}}}&{de}&{0.152}&{0.009}&{0.709}&{0.670}\\
    {}&es&{0.081}&{0.003}&{0.686}&{0.566}\\
    {}&fr&{0.124}&{0.007}&{0.688}&{0.650}\\
    {}&it&{0.084}&{0.005}&{0.604}&{0.562}\\
    {}&pt&{0.078}&{0.004}&{0.637}&{0.605}\\
    {}&ru&{0.129}&{0.006}&{0.712}&{0.678}\\
\hline
    \parbox[t]{2mm}{\multirow{5}{*}{\rotatebox[origin=c]{90}{WN}}}&{de}&{0.133}&{0.016}&{0.954}&{0.827}\\
    {}&es&{0.063}&{0.002}&{0.936}&{0.618}\\
    {}&fr&{0.114}&{0.012}&{0.938}&{0.799}\\
    {}&it&{0.085}&{0.009}&{0.961}&{0.785}\\
    {}&pt&{0.075}&{0.005}&{0.848}&{0.630}\\
    {}&ru&{0.167}&{0.024}&{0.982}&{0.890}\\
\hline
    \parbox[t]{2mm}{\multirow{5}{*}{\rotatebox[origin=c]{90}{WN-text}}}&{de}&{0.182}&{0.039}&{1.000}&{0.988}\\
    {}&es&{0.082}&{0.008}&{1.000}&{0.912}\\
    {}&fr&{0.144}&{0.030}&{1.000}&{0.981}\\
    {}&it&{0.111}&{0.021}&{1.000}&{0.963}\\
    {}&pt&{0.118}&{0.023}&{1.000}&{0.963}\\
    {}&ru&{0.192}&{0.037}&{1.000}&{0.991}\\
\hline
    \parbox[t]{2mm}{\multirow{5}{*}{\rotatebox[origin=c]{90}{Flores}}}&{de}&{0.084}&{0.002}&{0.709}&{0.502}\\
    {}&es&{0.066}&{0.000}&{0.914}&{0.502}\\
    {}&fr&{0.084}&{0.002}&{0.746}&{0.526}\\
    {}&it&{0.069}&{0.001}&{0.820}&\cellcolor{yellow!50}{0.494}\\   
    {}&pt&{0.053}&{0.001}&{0.713}&{0.502}\\
    {}&ru&{0.186}&{0.006}&{0.926}&{0.696}\\
\hline
\end{tabular}
\caption{Performance of the linear transform $e \rightarrow \tilde{e}$ (eq.\ref{eq:linear}), trained on Tatoeba dataset, and evaluated on (set aside) Tatoeba, WN (Wiki-news title-to-title), WN-text (Wiki-news title-to-halftext), and Flores. Performance is estimated as improvement in average distance $dD$ (eq.\ref{eq:estimate_by_dist}) and in average cosine $dC$ (eq.\ref{eq:estimate_by_cos}), fraction of samples with improved distance $fD$ (eq.\ref{eq:estimate_by_f_dist}), and fraction of samples with improved cosine $fC$ (eq.\ref{eq:estimate_by_f_cos}).}
\label{tab:eval_tatoeba}
\end{table}
The measure
\begin{equation} \label{eq:estimate_by_dist}
dD = \frac{d - \tilde{d}}{\min{(d, \tilde{d})}}
\end{equation}
compares the achieved average distance
\begin{equation} \label{eq:dist_avg_mapped}
\tilde{d} = \frac{1}{N}\sum_i^N |\tilde{e_i} - e'_i|
\end{equation}
and the original distance
\begin{equation} \label{eq:dist_avg_original}
d = \frac{1}{N}\sum_i^N |e_i - e'_i|
\end{equation}
where the embeddings $e$ are taken for a test dataset of size $N$.
The measure
\begin{equation} \label{eq:estimate_by_cos}
dC = \frac{1}{N}\sum_i^N \left(\cos(\tilde{e_i},e'_i) - \cos(e_i,e'_i)\right)
\end{equation}
compares the cosines. It is similar to comparing distances in eq.~\ref{eq:estimate_by_dist}; there is no need here for normalization, and the improvement is measured by the increase of cosine (whereas in eq.~\ref{eq:estimate_by_dist} it was the decrease of distance).

While on average the alignment of the embeddings may improve (as indeed is the case in our evaluations, showing $dD$ and $dC$ being positive in Table ~\ref{tab:eval_tatoeba}), the improvement is not evenly distributed between the samples. We would like to assess how many samples benefit from the transformation. The measure
\begin{equation} \label{eq:estimate_by_f_dist}
fD = \frac{1}{N}\sum_i^N \left(H(|e_i - e'_i| - |\tilde{e_i} - e'_i|)\right)
\end{equation}
where $H$ is the Heaviside step function, represents the fraction of the samples for which the distance has decreased.

Similarly, the measure
\begin{equation} \label{eq:estimate_by_f_cos}
\hspace{-3pt}fC = \frac{1}{N}\sum_i^N \left(H(\cos(\tilde{e_i},e'_i) - \cos(e_i,e'_i))\right)
\end{equation}
represents the fraction of the samples for which the cosine increased.

The transformation $e \rightarrow \tilde{e}$ helps if $dD$ and $dC$ are positive (the higher the better), and if the fractions $fD$ and $fC$ are higher than $0.5$ (the higher the better, for having an improvement in the majority of samples). The measures $dD$ and $fD$ should be of interest when matching of embeddings (e.g. search) is to be done by distance; the measures $dC$ and $fC$ are of interest for matching by cosine. 
Table \ref{tab:eval_tatoeba} shows that these conditions are satisfied for almost all cases. The only exception is the value of $fC$ for Italian ($it$) language in Flores dataset: here the cosine got improved for slightly less that half ($49.4\%$) of the samples.

\subsection{Orthogonality}\label{ssec:Linear_Orthogonality}

If a good translation indeed fully preserves the semantics of a sentence, and if the embedding model would produce ideal alignment, then the sentence embeddings in different languages would be close to identical: $e'=e$.
The transform $T$ (eq.\ref{eq:rotation}) would then become an identity. If the embedding model does not perfectly align the embeddings $e$ and $e'$ (or does not align them at all), but still correctly embed their semantics in each of the languages $L$ and $L'$, then the optimized linear transformation $(A,b)$ (eq.\ref{eq:linear}) must be orthogonal as in eq.\ref{eq:orthonormal_dilation_shift}.

In order to evaluate how close our linear transformation $A$ (trained on Tatoeba) to being orthogonal (Eq.\ref{eq:orthonormal}), we consider the values
\begin{equation} \label{eq:AA_nondiagonal}
p_{jk} = \frac{\sum_i{A_{ij}A_{ik}}}{|A_j|\cdot|A_k|} \hspace{4pt}, \hspace{16pt} j \neq k
\end{equation}
where 
\begin{equation} \label{eq:A_length}
|A_j| = \sqrt{\sum_iA_{ij}^2}
\end{equation}
The closer these values $p_{jk}$ to zero, the closer $A$ to being orthogonal.
In Table \ref{tab:eval_orthogonality_Tatoeba} we show simple aggregates of $p_{ij}$ over all $i \neq j$. The measure $\langle|p|\rangle$ is an average of absolute values of non-diagonal elements:
\begin{equation} \label{eq:avg_abs_p}
\langle|p|\rangle = \frac{1}{M(M-1)}\sum_{j\neq k}|p_{jk}|
\end{equation}
where $M$ is the dimensionality of the embeddings ($j,k=0,1, ..., M-1$).

The orthogonality may be compromised for some embeddings more than for others. To characterise this, we show in Table \ref{tab:eval_orthogonality_Tatoeba} the standard deviation 
\begin{equation} \label{eq:sigma_p}
\sigma(p) = \sqrt{\frac{1}{M(M-1)}\sum_{j\neq k}(p_{jk} - \langle p\rangle)^2}
\end{equation}
where the average $\langle p\rangle$ is
\begin{equation} \label{eq:avg_p}
\langle p\rangle = \frac{1}{M(M-1)}\sum_{j\neq k}p_{jk}
\end{equation}
We show also $\min(p)$ and $\max(p)$:
\begin{equation} \label{eq:p_min_max}
\min(p) = \min_{j\neq k}p_{jk} \:\quad \max(p) = \max_{j\neq k}p_{jk}
\end{equation}
Table \ref{tab:eval_orthogonality_Tatoeba} lists more languages than Table \ref{tab:eval_tatoeba} because there is no need here to apply $A$ to other datasets: we are simply considering the orthogonality of $A$. 

The highest by far deviation from orthogonality in Table \ref{tab:eval_orthogonality_Tatoeba} is for Berber ($ber$) language, followed by Esperanto ($eo$). 
The minimal and maximal values are colored yellow when they exceed $0.383$, meaning that for at least one pair $i,j$ the angle is less than $75\%$ of orthogonal ($\cos(\pi/2 * 0.75) \approx 0.383$.

\begin{table}[th!]
\centering
\begin{tabular}{@{}lcccc@{}}
\hline
{lang}&{$\langle|p|\rangle$}&{$\sigma(p)$}&{$\min(p)$}&{$\max(p)$}\\
\hline
{ber}&{0.204}&{0.254}&\cellcolor{yellow!50}{-0.861}&\cellcolor{yellow!50}{0.845}\\
{de}&{0.019}&{0.025}&{-0.154}&{0.337}\\
{eo}&{0.059}&{0.074}&{-0.362}&{0.345}\\
{es}&{0.004}&{0.005}&{-0.035}&{0.038}\\
{fr}&{0.019}&{0.024}&{-0.194}&\cellcolor{yellow!50}{0.397}\\
{he}&{0.027}&{0.034}&{-0.353}&\cellcolor{yellow!50}{0.516}\\
{it}&{0.011}&{0.014}&{-0.071}&{0.071}\\
{ja}&{0.032}&{0.042}&{-0.360}&\cellcolor{yellow!50}{0.623}\\
{pt}&{0.013}&{0.017}&{-0.100}&{0.135}\\
{ru}&{0.018}&{0.023}&{-0.150}&{0.219}\\
{tr}&{0.027}&{0.035}&{-0.322}&\cellcolor{yellow!50}{0.498}\\
{uk}&{0.020}&{0.026}&{-0.191}&{0.281}\\
\hline
\end{tabular}
\caption{Aggregates over orthogonality conditions Eq.\ref{eq:AA_nondiagonal} for $A$ trained on Tatoeba dataset, for languages containing at least $100K$ samples. Min and max beyond $25\%$ deviation from orthogonality ($\cos(0.75\pi/2) \approx 0.383$) are colored yellow.}
\label{tab:eval_orthogonality_Tatoeba}
\end{table}

For comparison, in Table \ref{tab:eval_orthogonality_UN} we show similar data for $A$ trained on United Nations Parallel Corpus UNPC \cite{ziemski-etal-2016-united}\footnote{https://conferences.unite.un.org/uncorpus} (with 500K samples used for training and 10K for validation). The UN texts have a specific formal style and meant to be precise in dealing with loaded topics. The translations are also intended to be precise, conserving semantics. But these documents' cumbersome formal style and some very long sentences may be more difficult than the common texts for an embedding model. 
Indeed, for each of the three languages common for Tatoeba Table \ref{tab:eval_orthogonality_Tatoeba} and UNPC Table \ref{tab:eval_orthogonality_UN} (Spanish $es$, French $fr$ and Russian $ru$) all the aggregate indicators $\langle|p|\rangle$, $\sigma(p)$, $\min(p)$ and $\max(p)$ are several times larger for UNPC-trained matrix $A$ (Table~\ref{tab:eval_orthogonality_UN}).  

\begin{table}[th!]
\centering
\begin{tabular}{@{}lcccc@{}}
\hline
{lang}&{$\langle|p|\rangle$}&{$\sigma(p)$}&{$\min(p)$}&{$\max(p)$}\\
\hline
{ar}&{0.026}&{0.033}&{-0.147}&{0.157}\\
{es}&{0.014}&{0.018}&{-0.130}&{0.107}\\
{fr}&{0.144}&{0.195}&\cellcolor{yellow!50}{-0.769}&\cellcolor{yellow!50}{0.795}\\
{ru}&{0.404}&{0.476}&\cellcolor{yellow!50}{-0.958}&\cellcolor{yellow!50}{0.950}\\
{zh}&{0.039}&{0.050}&{-0.254}&\cellcolor{yellow!50}{0.495}\\
\hline
\end{tabular}
\caption{Aggregates over orthogonality conditions Eq.\ref{eq:AA_nondiagonal} for $A$ trained on UNPC.}
\label{tab:eval_orthogonality_UN}
\end{table}

The orthogonal transformation can be accompanied by dilation (coefficient $\alpha$ in Eq.\ref{eq:orthonormal_dilation_shift}), which means that the values $\alpha_i = |A_i|$ (eq.\ref{eq:A_length}) should not depend on $i$. 
In order to assess deviations from this condition, we consider normalized standard deviation
\begin{equation} \label{eq:stretch_nstdev}
\frac{\sigma(\alpha)}{\bar\alpha} = \frac{1}{\bar{\alpha}}\sqrt{\frac{1}{M}\sum_i(\alpha_i - \bar{\alpha})^2}
\end{equation}
and normalized range 
\begin{equation} \label{eq:stretch_range}
r(\alpha) = \frac{\max{\alpha} - \min{\alpha}}{\bar{\alpha}}
\end{equation}
where
\begin{equation} \label{eq:alpha_avg}
\bar{\alpha} = \frac{1}{M}\sum_{i}\alpha_i
\end{equation}
\begin{equation} \label{eq:alpha_min_max}
\min(\alpha) = \min_{i}\alpha_i \:\quad \max(\alpha) = \max_{i}\alpha_i
\end{equation}

The dilation quality measures $\frac{\sigma(\alpha)}{\bar\alpha}$ and $r(\alpha)$ are shown in Tables \ref{tab:eval_orthogonality_stretch_Tatoeba} and \ref{tab:eval_orthogonality_stretch_UN}, for the transformations obtained on Tatoeba and on UNPC datasets correspondingly. The tables contain also the values of $\bar{\alpha}$ - the averaged $\alpha$, and of the minimal and maximal values of $\alpha$.

\begin{table}[th!]
\centering
\begin{tabular}{@{}lccccc@{}}
\hline
{lang}&{$\bar{\alpha}$}&{$\frac{\sigma(\alpha)}{\bar\alpha}$}&{$r(\alpha)$}&{$\min(\alpha)$}&{$\max(\alpha)$}\\
\hline
{ber}&{0.637}&{0.336}&{1.856}&{0.258}&{1.440}\\
{de}&{0.814}&{0.039}&{0.275}&{0.753}&{0.977}\\
{eo}&{0.640}&{0.192}&{1.056}&{0.377}&{1.053}\\
{es}&{0.964}&{0.005}&{0.050}&{0.951}&{1.000}\\
{fr}&{0.845}&{0.034}&{0.230}&{0.791}&{0.986}\\
{he}&{0.814}&{0.060}&{0.333}&{0.727}&{0.998}\\
{it}&{0.889}&{0.021}&{0.170}&{0.841}&{0.992}\\
{ja}&{0.809}&{0.073}&{0.419}&{0.705}&{1.044}\\
{pt}&{0.877}&{0.022}&{0.174}&{0.838}&{0.990}\\
{ru}&{0.836}&{0.031}&{0.224}&{0.789}&{0.976}\\
{tr}&{0.835}&{0.054}&{0.307}&{0.751}&{1.007}\\
{uk}&{0.860}&{0.037}&{0.239}&{0.797}&{1.002}\\
\hline
\end{tabular}
\caption{Nonuniformity of dilation of embeddings transformation (Eqs.\ref{eq:stretch_nstdev}, \ref{eq:stretch_range}). For the transformation trained on Tatoeba dataset.}
\label{tab:eval_orthogonality_stretch_Tatoeba}
\end{table}

\begin{table}[th!]
\centering
\begin{tabular}{@{}lccccc@{}}
\hline
{lang}&{$\bar{\alpha}$}&{$\frac{\sigma(\alpha)}{\bar\alpha}$}&{$r(\alpha)$}&{$\min(\alpha)$}&{$\max(\alpha)$}\\
\hline
{ar}&{0.761}&{0.088}&{0.470}&{0.630}&{0.988}\\
{es}&{0.840}&{0.043}&{0.253}&{0.767}&{0.980}\\
{fr}&{0.938}&{0.190}&{1.126}&{0.700}&{1.756}\\
{ru}&{1.338}&{0.444}&{2.908}&{0.661}&{4.551}\\
{zh}&{0.865}&{0.114}&{0.559}&{0.696}&{1.180}\\
\hline
\end{tabular}
\caption{Nonuniformity of dilation of embeddings transformation (Eqs.\ref{eq:stretch_nstdev}, \ref{eq:stretch_range}). For the transformation trained on UNPC.}
\label{tab:eval_orthogonality_stretch_UN}
\end{table}

Similarly to the orthogonality conditions, the dilation quality measures $\frac{\sigma(\alpha)}{\bar\alpha}$ and $r(\alpha)$ are better (lower) for the transformation trained on Tatoeba (Table \ref{tab:eval_orthogonality_stretch_Tatoeba}) than on UNPC (Table \ref{tab:eval_orthogonality_stretch_UN}), for all three languages they have in common: Spanish ($es$), French ($fr$) and Russian ($ru$).
Both measures generally follow similar trends across the languages.

As we could already expect from observations in Table ~\ref{tab:eval_orthogonality_Tatoeba}, the measures $\frac{\sigma(\alpha)}{\bar\alpha}$ and $r(\alpha)$ in Table ~\ref{tab:eval_orthogonality_stretch_Tatoeba} are the worst for Berber ($ber$) and Esperanto ($eo$) languages. A distant third (also as in Table ~\ref{tab:eval_orthogonality_stretch_Tatoeba}) is Japanese language ($ja$).

For most languages the ratio $\frac{\sigma(\alpha)}{\bar\alpha}$ may look comfortably small, but the normalized range $r(\alpha)$ is high for some languages in both tables \ref{tab:eval_orthogonality_stretch_Tatoeba} and \ref{tab:eval_orthogonality_stretch_UN}.
Altogether, we have to conclude that orthogonality is only approximately satisfied by the linear transform $(A,b)$.

\section{Conclusion}\label{sec:Conclusion}
We considered a simple and inexpensive method of improving the alignment between sentence embeddings in two languages: a linear transformation, tuned on embeddings of the paired sentences. In the examples we analyzed, a training on sentences also improves an alignment between titles and texts (lower-half texts) of the articles - the articles from our WikiNews dataset.

If embeddings were capable of perfectly encoding semantics even when not perfectly aligned, then the linear transformation would be an orthogonal transformation, accompanied by dilation and shift. Measuring deviation from this condition allows us to judge the quality of the embeddings. For example, we observed lower quality for embeddings of Berber and Esperanto languages compared to other languages considered here, and also a lower quality of UNPC-trained transformations compared to Tatoeba-trained transformations.

It would be interesting to consider deviation from orthogonality for individual samples, as the strong deviations could point either to bad translations or to the samples difficult to embed by the model.

\section*{Limitations}\label{sec:Limitations}
Our consideration involved a limited set of languages. This limitation allowed us to evaluate Tatoeba-trained transformations on very different styles of matching sentences, but the research can be extended to many more languages. 

We suggested simple measures of quality of multilingual embeddings based on the orthogonality requirement (Section ~\ref{ssec:Linear_Orthogonality}). While our observations confirm that these measures are reasonable, we do not claim that these are the best possible measures.  

We have not considered a possibility of measuring the deviations from orthogonality by individual samples. If such samples are particularly imperfect translation (see Appendix ~\ref{app:ambiguity_loss}) then removing such samples from the dataset used for tuning would improve orthogonality of the transformation, and hence would make better the introduced here measures of the quality of embeddings. 

A complementing possibility is that a very good embedding model could help to identify imperfect translations; this may be unlikely because the embeddings are very approximate in encoding the semantics, but we do not provide definitive observations.

The role of polysemy and its variation between languages is not investigated here beyond the intuitive arguments and examples of Appendix ~\ref{app:ambiguity_loss}, in which we suggest that in most cases the context removes or strongly reduces ambiguity, and a good translation keeps the residual ambiguity, if any, unchanged.

\section*{Acknowledgments}
We thank Randy Sawaya for many discussions and review of the paper. We also thank an anonymous reviewer for concern about the polysemy problem.


\bibliography{custom}

\begin{thebibliography}{16}
\expandafter\ifx\csname natexlab\endcsname\relax\def\natexlab#1{#1}\fi

\bibitem[{Cao et~al.(2020)Cao, Kitaev, and Klein}]{Cao2020MultilingualAlignment}
Steven Cao, Nikita Kitaev, and Dan Klein. 2020.
\newblock \href {https://doi.org/10.48550/ARXIV.2002.03518} {Multilingual alignment of contextual word representations}.
\newblock \emph{arXiv}, arXiv:2002.03518.

\bibitem[{Casas et~al.(2019)Casas, Hernández-Fernández, Català, i~Cancho, and Baixeries}]{CASAS201919}
Bernardino Casas, Antoni Hernández-Fernández, Neus Català, Ramon~Ferrer i~Cancho, and Jaume Baixeries. 2019.
\newblock \href {https://doi.org/https://doi.org/10.1016/j.csl.2019.03.007} {Polysemy and brevity versus frequency in language}.
\newblock \emph{Computer Speech \& Language}, 58:19--50.

\bibitem[{Goyal et~al.(2022)Goyal, Gao, Chaudhary, Chen, Wenzek, Ju, Krishnan, Ranzato, Guzm{\'a}n, and Fan}]{goyal-etal-2022-flores}
Naman Goyal, Cynthia Gao, Vishrav Chaudhary, Peng-Jen Chen, Guillaume Wenzek, Da~Ju, Sanjana Krishnan, Marc{'}Aurelio Ranzato, Francisco Guzm{\'a}n, and Angela Fan. 2022.
\newblock \href {https://doi.org/10.1162/tacl_a_00474} {The {F}lores-101 evaluation benchmark for low-resource and multilingual machine translation}.
\newblock \emph{Transactions of the Association for Computational Linguistics}, 10:522--538.

\bibitem[{Guzm{\'a}n et~al.(2019)Guzm{\'a}n, Chen, Ott, Pino, Lample, Koehn, Chaudhary, and Ranzato}]{guzman-etal-2019-flores}
Francisco Guzm{\'a}n, Peng-Jen Chen, Myle Ott, Juan Pino, Guillaume Lample, Philipp Koehn, Vishrav Chaudhary, and Marc{'}Aurelio Ranzato. 2019.
\newblock \href {https://doi.org/10.18653/v1/D19-1632} {The {FLORES} evaluation datasets for low-resource machine translation: {N}epali{--}{E}nglish and {S}inhala{--}{E}nglish}.
\newblock In \emph{Proceedings of the 2019 Conference on Empirical Methods in Natural Language Processing and the 9th International Joint Conference on Natural Language Processing (EMNLP-IJCNLP)}, pages 6098--6111, Hong Kong, China. Association for Computational Linguistics.

\bibitem[{Kang et~al.(2024)Kang, Blevins, and Zettlemoyer}]{kang-etal-2024-translate}
Haoqiang Kang, Terra Blevins, and Luke Zettlemoyer. 2024.
\newblock \href {https://aclanthology.org/2024.eacl-long.94} {Translate to disambiguate: Zero-shot multilingual word sense disambiguation with pretrained language models}.
\newblock In \emph{Proceedings of the 18th Conference of the European Chapter of the Association for Computational Linguistics (Volume 1: Long Papers)}, pages 1562--1575, St. Julian{'}s, Malta. Association for Computational Linguistics.

\bibitem[{Mikolov et~al.(2013)Mikolov, Le, and Sutskever}]{https://doi.org/10.48550/arxiv.1309.4168}
Tomas Mikolov, Quoc~V. Le, and Ilya Sutskever. 2013.
\newblock \href {https://doi.org/10.48550/ARXIV.1309.4168} {Exploiting similarities among languages for machine translation}.
\newblock \emph{arXiv}, arXiv:1309.4168.

\bibitem[{Patra et~al.(2019)Patra, Moniz, Garg, Gormley, and Neubig}]{patra-etal-2019-bilingual}
Barun Patra, Joel Ruben~Antony Moniz, Sarthak Garg, Matthew~R. Gormley, and Graham Neubig. 2019.
\newblock \href {https://doi.org/10.18653/v1/P19-1018} {Bilingual lexicon induction with semi-supervision in non-isometric embedding spaces}.
\newblock In \emph{Proceedings of the 57th Annual Meeting of the Association for Computational Linguistics}, pages 184--193, Florence, Italy. Association for Computational Linguistics.

\bibitem[{Peng et~al.(2020)Peng, Stevenson, Lin, and Li}]{Peng2022Understanding}
Xutan Peng, Mark Stevenson, Chenghua Lin, and Chen Li. 2020.
\newblock \href {https://doi.org/10.48550/ARXIV.2004.01079} {Understanding linearity of cross-lingual word embedding mappings}.
\newblock \emph{arXiv}, arXiv:2004.01079.

\bibitem[{Reimers and Gurevych(2019)}]{reimers-2019-sentence-bert}
Nils Reimers and Iryna Gurevych. 2019.
\newblock \href {http://arxiv.org/abs/1908.10084} {Sentence-bert: Sentence embeddings using siamese bert-networks}.
\newblock In \emph{Proceedings of the 2019 Conference on Empirical Methods in Natural Language Processing}. Association for Computational Linguistics.

\bibitem[{Reimers and Gurevych(2020)}]{reimers-gurevych-2020-making}
Nils Reimers and Iryna Gurevych. 2020.
\newblock \href {https://doi.org/10.18653/v1/2020.emnlp-main.365} {Making monolingual sentence embeddings multilingual using knowledge distillation}.
\newblock In \emph{Proceedings of the 2020 Conference on Empirical Methods in Natural Language Processing (EMNLP)}, pages 4512--4525, Online. Association for Computational Linguistics.

\bibitem[{Srinivasan and Rabagliati(2015)}]{SRINIVASAN2015124}
Mahesh Srinivasan and Hugh Rabagliati. 2015.
\newblock \href {https://doi.org/https://doi.org/10.1016/j.lingua.2014.12.004} {How concepts and conventions structure the lexicon: Cross-linguistic evidence from polysemy}.
\newblock \emph{Lingua}, 157:124--152.
\newblock Polysemy: Current Perspectives and Approaches.

\bibitem[{Team et~al.(2022)Team, Costa-jussà, Cross, Çelebi, Elbayad, Heafield, Heffernan, Kalbassi, Lam, Licht, Maillard, Sun, Wang, Wenzek, Youngblood, Akula, Barrault, Gonzalez, Hansanti, Hoffman, Jarrett, Sadagopan, Rowe, Spruit, Tran, Andrews, Ayan, Bhosale, Edunov, Fan, Gao, Goswami, Guzmán, Koehn, Mourachko, Ropers, Saleem, Schwenk, and Wang}]{nllbteam2022language}
NLLB Team, Marta~R. Costa-jussà, James Cross, Onur Çelebi, Maha Elbayad, Kenneth Heafield, Kevin Heffernan, Elahe Kalbassi, Janice Lam, Daniel Licht, Jean Maillard, Anna Sun, Skyler Wang, Guillaume Wenzek, Al~Youngblood, Bapi Akula, Loic Barrault, Gabriel~Mejia Gonzalez, Prangthip Hansanti, John Hoffman, Semarley Jarrett, Kaushik~Ram Sadagopan, Dirk Rowe, Shannon Spruit, Chau Tran, Pierre Andrews, Necip~Fazil Ayan, Shruti Bhosale, Sergey Edunov, Angela Fan, Cynthia Gao, Vedanuj Goswami, Francisco Guzmán, Philipp Koehn, Alexandre Mourachko, Christophe Ropers, Safiyyah Saleem, Holger Schwenk, and Jeff Wang. 2022.
\newblock \href {http://arxiv.org/abs/2207.04672} {No language left behind: Scaling human-centered machine translation}.
\newblock \emph{arXiv}, arXiv:2207.04672.

\bibitem[{Xie et~al.(2022)Xie, Zhao, Yu, and Li}]{xie-etal-2022-discovering}
Zhihui Xie, Handong Zhao, Tong Yu, and Shuai Li. 2022.
\newblock \href {https://aclanthology.org/2022.emnlp-main.379} {Discovering low-rank subspaces for language-agnostic multilingual representations}.
\newblock In \emph{Proceedings of the 2022 Conference on Empirical Methods in Natural Language Processing}, pages 5617--5633, Abu Dhabi, United Arab Emirates. Association for Computational Linguistics.

\bibitem[{Yang et~al.(2021)Yang, Yang, Cer, and Darve}]{yang-etal-2021-simple}
Ziyi Yang, Yinfei Yang, Daniel Cer, and Eric Darve. 2021.
\newblock \href {https://doi.org/10.18653/v1/2021.emnlp-main.470} {A simple and effective method to eliminate the self language bias in multilingual representations}.
\newblock In \emph{Proceedings of the 2021 Conference on Empirical Methods in Natural Language Processing}, pages 5825--5832, Online and Punta Cana, Dominican Republic. Association for Computational Linguistics.

\bibitem[{Zhao and Gilman(2020)}]{zhao-gilman-2020-non}
Jiawei Zhao and Andrew Gilman. 2020.
\newblock \href {https://aclanthology.org/2020.lrec-1.440} {Non-linearity in mapping based cross-lingual word embeddings}.
\newblock In \emph{Proceedings of the Twelfth Language Resources and Evaluation Conference}, pages 3583--3589, Marseille, France. European Language Resources Association.

\bibitem[{Ziemski et~al.(2016)Ziemski, Junczys-Dowmunt, and Pouliquen}]{ziemski-etal-2016-united}
Micha{\l} Ziemski, Marcin Junczys-Dowmunt, and Bruno Pouliquen. 2016.
\newblock \href {https://aclanthology.org/L16-1561} {The {U}nited {N}ations parallel corpus v1.0}.
\newblock In \emph{Proceedings of the Tenth International Conference on Language Resources and Evaluation ({LREC}'16)}, pages 3530--3534, Portoro{\v{z}}, Slovenia. European Language Resources Association (ELRA).

\end{thebibliography}

\appendix

\section{Loss of Ambiguity}
\label{app:ambiguity_loss}
\subsection{Polysemy problem}
\label{app:polysemy}
As we discussed in Introduction (Section ~\ref{sec:Intro}), we assume that ambiguity of words is mostly lost with context. 
On one hand, it is intuitively understandable, and the role of context was demonstrated in \cite{kang-etal-2024-translate}. 
On the other hand, polysemy is occasionally possible even with a context. A word ambiguity can be intentionally and skillfully kept through many sentences or a long dialog for sake of misinterpretation comedy. Also, the available word polysemy greatly varies across languages and patterns (see for example Table 5 in \cite{SRINIVASAN2015124}, or Table 3 in \cite{CASAS201919}).

Yet, the common sentences are remarkably unambiguous. For example, the word `board' looses all or most of its multiple definitions in arbitrary sentences of lengths 3 to 7 words, generated by a GPT3.5 (our examples are in Table ~\ref{tab:sentences_board}).  

\begin{table}[th!]
\centering
\begin{tabular}{@{}l@{}}
\hline
{\small The board cracked.}\\
{\small The board is white.}\\
{\small The board is now full.}\\
{\small Circuit board malfunctioned, causing system failure.}\\
{\small The cork board holds important reminders daily.}\\
\hline
\end{tabular}
\caption{Examples of sentences with the word `board'.}
\label{tab:sentences_board}
\end{table}

Even the residual ambiguity may be kept unchanged by a good translation. As a simple illustration that a typical sentence is loosing ambiguity of its words, and that the residual ambiguity is usually kept intact in a good translation, we examined (in the following subsections) the first 10 sentences from Tatoeba and from Flores datasets, reviewing the sentences in English, French, Japanese, Russian, Spanish and Ukrainian. (We also reviewed top 10 sentences from UNPS in English, French, Russian and Spanish, and could not find any change in semantics in those meticulous formal style translations.)

Despite many multi-sense words in all the sentences considered below, there were few examples where the semantics of English sentence would not exactly correspond to semantics of the translated sentence.
The examples show that if semantics of a sentence is somewhat changed in translation, it is mostly due to a deficiency of translation rather than some impenetrable polysemy barrier between the languages.

\subsection{Examples from Tatoeba}
\label{ssec:ambiguity_tatoeba}
We have not found any shift of semantics in the first 10 samples from English-French part of Tatoeba. For example, the very first sample uses a few words that could be used in different senses, but the semantics of English \foreignlanguage{english}{"When he asked who had broken the window, all the boys put on an air of innocence."} is well matched by French "Lorsqu'il a demandé qui avait cassé la fenêtre, tous les garçons ont pris un air innocent.".

In the case of the first 10 samples of English-Japanese pairs from Tatoeba, one example has a shift in semantics. The Japanese translation "\begin{CJK}{UTF8}{min}ムーリエルは２０歳になりました。\end{CJK}" means Muiriel has just turned 20 years old, whereas the English translation "Muiriel is 20 now." does not indicate whether Muiriel has just turned 20 or has been 20 for a while.
Additionally, the over-restriction of the meaning of "20" to age, as seen on English-Ukrainian pairs, is also observed in the translation to Japanese.

Of the first 10 samples from English-Spanish part of Tatoeba, we found 2 samples where the semantics is shifted: The English sentence \foreignlanguage{english}{"Let's try something."} is translated (in two out of four versions) using the word \foreignlanguage{english}{"permiteme"}, which narrows down the meaning by suggesting that  it is the speaker that would \foreignlanguage{english}{"try something"}.

Of the first 10 samples from English-Ukrainian part of Tatoeba, we found 2 samples where semantics is somewhat shifted. One of three translations of the English sentence \foreignlanguage{english}{"I have to go to sleep."} is over-specific \foreignlanguage{ukrainian}{\enquote{Мені час йти спати.}}, narrowing the reason (time). Also, one of three translations of \foreignlanguage{english}{"Muiriel is 20 now."} is \foreignlanguage{russian}{"Мюріел зараз двадцять років.".} Strictly speaking, the English sentence could also be used in a game or sport to inform about some score, while this particular translation to Ukrainian narrows down \foreignlanguage{english}{"20"} as age. 

Of the first 10 samples from English-Russian part of Tatoeba, there is one sample with shifted semantics: Similar to Ukrainian samples, one of three translations of \foreignlanguage{english}{"I have to go to sleep."} is over-specific \foreignlanguage{russian}{"Мне пора идти спать."} (meaning "It is time for me to go to sleep.").

\subsection{Examples from Flores}
\label{ssec:ambiguity_flores}
Sentences in Flores, unlike in Tatoeba, are long sentences like one would encounter in informative news. Each sample consists of an English sentence is translated to many languages. Of the first 10 samples examined for translation to French, Japanese, Russian, Spanish and Ukrainian languages, we could find only three examples of the translation changing semantics.

There are two examples in English-Japanese pairs where the Japanese translations differ semantically from the English ones. In sample \#1, the English phrase "about one U.S. cent each" is translated to "\begin{CJK}{UTF8}{min}1円ほどす。\end{CJK}". First, there is a typographical error where "\begin{CJK}{UTF8}{min}ほどす。\end{CJK}" should be typed "\begin{CJK}{UTF8}{min}ほどです。\end{CJK}" . This could result in misalignment of the semantics of the sentence pair. Secondly, there is a semantic shift in translation. Its literal translation is "about 1 yen", which uses Japanese currency. Although a cent and a yen are of similar value (currently, 1 cent is about 1.6 yen), changing the currency unit in translation can significantly alter the sentence's meaning. Another example is the English phrase "closing the airport to commercial flights" in sample \#3. Its Japanese translation is "\begin{CJK}{UTF8}{min}空港の商業便が閉鎖されました。\end{CJK}", which literally means "Commercial flights in the airport were closed," where the object of the verb "close" is "flights," not the airport. Confusing the subject and object can change the semantic meaning of the sentences.

There is one example (sample \#8) of semantics shift in English-Ukrainian pairs: In the translation of English sentence \foreignlanguage{english}{"The protest started around 11:00 local time (UTC+1) on Whitehall opposite the police-guarded entrance to Downing Street, the Prime Minister's official residence."} to Ukrainian the word \foreignlanguage{russian}{"біля"} (meaning "near") was used for \foreignlanguage{english}{"opposite"}, thus adding a bit of ambiguity.

\section{WikiNews}
\label{app:wikinews}

\begin{table*}[]
\centering
\caption{Example of samples from the multilingual WikiNews dataset}
\begin{tabular}{@{}cllll@{}}
\toprule
index & \textbf{pageid} & \textbf{lang} & \textbf{title} & \textbf{content} \\
\midrule
0 & 232226 & en & \multirow{2}{*}{\parbox{5cm}{``Very serious'': Chinese government releases corruption report}} & \multirow{2}{*}{\parbox{6cm}{A report by the Chinese government states corruption is "very serious". ...}} \\
 & & & & \\
1 & 232226 & cs & \multirow{2}{*}{\parbox{5cm}{Čína připustila, že tamní \\ korupce je vážný problém}} & \multirow{2}{*}{\parbox{7cm}{Zpráva čínské vlády připouští, že korupce v zemi je stále „velmi vážná“, jelikož úřady ...}} \\
 & & & & \\
2 & 232226 & es & \multirow{3}{*}{\parbox{5cm}{China admite que la corrupción \\ en el país es ``muy seria''s}} & \multirow{3}{*}{\parbox{7cm}{29 de diciembre de 2010Beijing, China — Un reporte del gobierno de la República Popular China ...}} \\
 & & & & \\
 & & & & \\
\bottomrule
\end{tabular}
\label{tab:example_wikinews}
\end{table*}

 The WikiNews dataset\footnote{https://huggingface.co//datasets//Fumika//Wikinews-multilingual}\footnote{https://github.com//PrimerAI//primer-research} comprises 15,200 news articles from the multilingual WikiNews website\footnote{https://www.wikinews.org/}, including 9,960 non-English articles written in 33 different languages. These articles are linked to one of 5,240 sets of English news articles as WikiNews pages in other languages. Therefore, these WikiPages in different languages can be assumed to be describing the same news event, thus we can assume that the news titles and contents are of the linked NewsPages are semantically alligned. Here the non-English articles are written in a variety of languages including Spanish, French, German, Portuguese, Polish, Italian, Chinese, Russian, Japanese, Dutch, Swedish, Tamil, Serbian, Czech, Catalan, Hebrew, Turkish, Finnish, Esperanto, Greek, Hungarian, Ukrainian, Norwegian, Arabic, Persian, Korean, Romanian, Bulgarian, Bosnian, Limburgish, Albanian, and Thai.

Each sample in the multilingual WikiNews dataset includes several variables, such as pageid, title, categories, language, URL, article content, and the publish date. In some cases, foreign WikiNews sites may have news titles but no content, in which case the text variable is left empty. Samples with the same pageid in the dataset correspond to the same news event, which are linked together as the same WikiNews pages with other languages. The published date of an English sample is scraped and converted to DateTime format, but dates in foreign samples are left as is.
Table \ref{tab:example_wikinews} shows the example samples of the dataset.

The number of samples for the languages used in Table \ref{tab:eval_tatoeba}: $de$: 1053; $es$: 1439; $fr$: 1311; $it$: 618; $pt$: 1023; $ru$: 436.

\end{document}